\documentclass[times, 10pt,twocolumn]{article} 
\newcommand{\final}{1}
\usepackage{latex8}
\usepackage{times}

\usepackage{multirow}
\usepackage{amsmath}
\usepackage{booktabs} 
\usepackage{wasysym}
\usepackage{graphicx} 
\usepackage{dsfont}
\usepackage{bbm}
\usepackage{array,color}
\usepackage{subfigure}
\usepackage{verbatim}
\usepackage{amssymb} 
\usepackage{ifthen}
\usepackage{url}
\usepackage{microtype}

\newcommand{\warning}[1]{{\it\color{red} #1}}
\newcommand{\toremove}[1]{{\it\color{red} (To remove) #1}}
\newcommand{\nothing}[1]{}

\definecolor{WeimingColor}{rgb}{0,0,0.8}
\newcommand{\weiming}[1]{{\color{WeimingColor} [Weiming: #1]}}
\definecolor{SongyuColor}{rgb}{0.8,0,0}
\newcommand{\songyu}[1]{{\color{SongyuColor} [Songyu: #1]}}
\definecolor{FanColor}{rgb}{0.8,0,0.8}
\newcommand{\fan}[1]{{\color{FanColor}[Fan: #1]}}
\definecolor{OliverColor}{rgb}{0.54,0.18,0.88}

\definecolor{AnswerColor}{rgb}{0.118,0.0667,0.418}

\definecolor{AudioColor}{rgb}{0.56,0.34,0.62}

\definecolor{figred}{rgb}{1,0,0}
\definecolor{figgreen}{rgb}{0,0.6,0}
\definecolor{figblue}{rgb}{0,0,1}
\definecolor{figpink}{rgb}{1,0.63,0.63}

\ifthenelse{\equal{\final}{1}}
{
\renewcommand{\weiming}[1]{}
\renewcommand{\songyu}[1]{}
\renewcommand{\fan}[1]{}
\renewcommand{\warning}[1]{}
\renewcommand{\toremove}[1]{}
}
{}

\begin{document}
\newcommand{\contentimage}{x}
\newcommand{\styleimage}{y}
\newcommand{\stylizedimage}{\tilde{y}}
\newcommand{\reconsimage}{\hat{x}}
\newcommand{\attribute}{c}
\newcommand{\artist}{a}
\newcommand{\genre}{g}
\newcommand{\period}{p}
\newcommand{\contentdomain}{\mathcal{X}}
\newcommand{\styledomain}{\mathcal{Y}}
\newcommand{\gaussiandistributionwithparams}{\mathcal{N}(\mu,\sigma^2)}
\newcommand{\forwardgen}{G}
\newcommand{\backwardgen}{F}
\newcommand{\forwarddis}{D_{y}}
\newcommand{\backwarddis}{D_{x}}
\newcommand{\mathenv}[1]{$#1$}
\newcommand{\belongto}[2]{\mathenv{#1\in#2}}

\title{Multi-Attribute Guided Painting Generation}

\author{Minxuan Lin\\
NLPR, CASIA \& School of AI, UCAS \\ linminxuan2018@ia.ac.cn\\
\and
Yingying Deng\\
School of AI, UCAS \& NLPR, CASIA\\
dengyingying2017@ia.ac.cn\\
\and
Fan Tang\\
Fosafer\\
tangfan@fosafer.com\\
\and
Weiming Dong~\thanks{Corresponding author}\\
NLPR, CASIA \& CASIA-LLVision Joint Lab\\
weiming.dong@ia.ac.cn\\
\and
Changsheng Xu\\
NLPR, CASIA\\
csxu@nlpr.ia.ac.cn\\
}

\maketitle
\thispagestyle{empty}

\begin{abstract}
Controllable painting generation plays a pivotal role in image stylization. 
Currently, the control way of style transfer is subject to exemplar-based reference or a random one-hot vector guidance. 
Few works focus on decoupling the intrinsic properties of painting as control conditions, e.g., artist, genre and period. 
Under this circumstance, we propose a novel framework adopting multiple attributes from the painting to control the stylized results. 
An asymmetrical cycle structure is equipped to preserve the fidelity, associating with style preserving and attribute regression loss to keep the unique distinction of colors and textures between domains. 
Several qualitative and quantitative results demonstrate the effect of the combinations of multiple attributes and achieve satisfactory performance.
\end{abstract}
\section{Introduction}
\begin{figure}[t]
    \centering
    \includegraphics[width=0.49\textwidth]{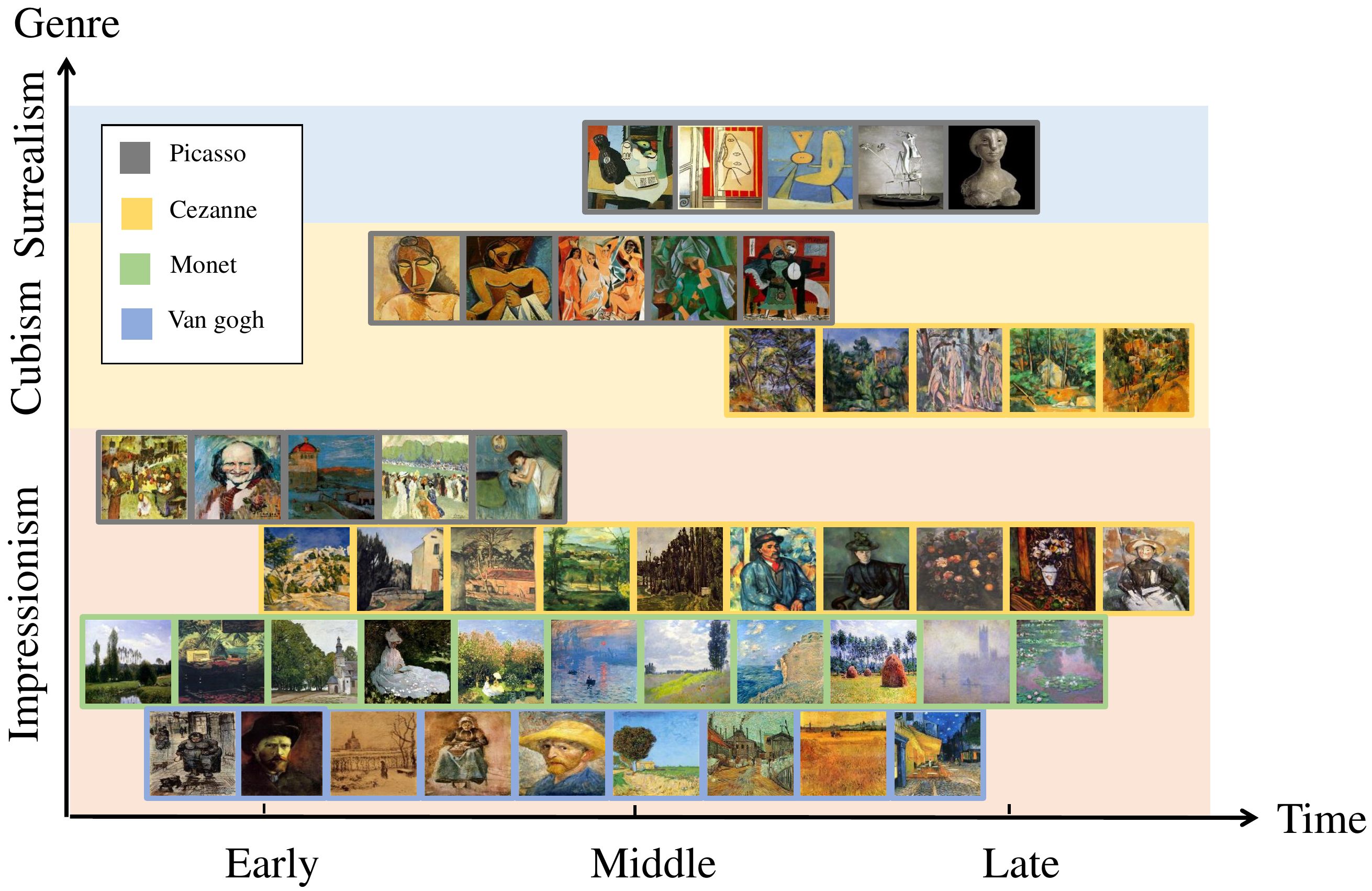}
    \caption{The artwork stream of diverse artists. The drawing habits of color and brush stroke gradually alter over period and genre. }
    \label{fig:art_history}
\end{figure}

Painting generation is a technique used to create art by synthesizing style patterns from given painting image(s) evenly over a natural image while maintaining its original content structure.
For art creation, every painter has a unique interpretation of painting style.
Moreover, the painting preferences (e.g., the usage of strokes and colors) change at different periods.
Commonly, the attributes like \emph{artist, genre} and \emph{period} are regarded as crucial factors to represent paintings, which will also affect the visual appearance of style transfer results.
Figure~\ref{fig:art_history} shows the art stream of individual painting preference along with genres and career periods. 
The disparities of colors and brush strokes reflect in different genres and periods, forming some representative features which belong to specific attribute combinations. 
Thus, using these attributes as control conditions is an acceptable measure to guide painting generation. 
However, most existing methods~\cite{park2019arbitrary,sanakoyeu:2018:style} mainly focus on increasing the quality of stylized results but are lack of high controllability in appearance guided by these properties.

To address this problem, we propose a practical approach, \textit{multi-attribute guided painting generation}, for artistic stylized image generation where painting appearance can be easily controlled by user-assigned attributes.
Specifically, motivated by CycleGAN~\cite{Zhu:2017:ICCV}, an \textit{asymmetrical} cycle structure is adopted as content branch to generate stylized results.
With regard to the multi-attribute as style control conditions, rather than simply stacking the conditions into the feature layers, a congregate multi-attribute vector is parsed by a multi-layer preception network as style guidance to produce AdaIN~\cite{Huang:2017:ICCV} parameters. 
To enhance the distinction, we utilize a multi-task discriminator and style preserving constraint to enlarge the color and texture gaps among different domains.

\begin{figure*}[t]
    \centering
    \includegraphics[width=\textwidth]{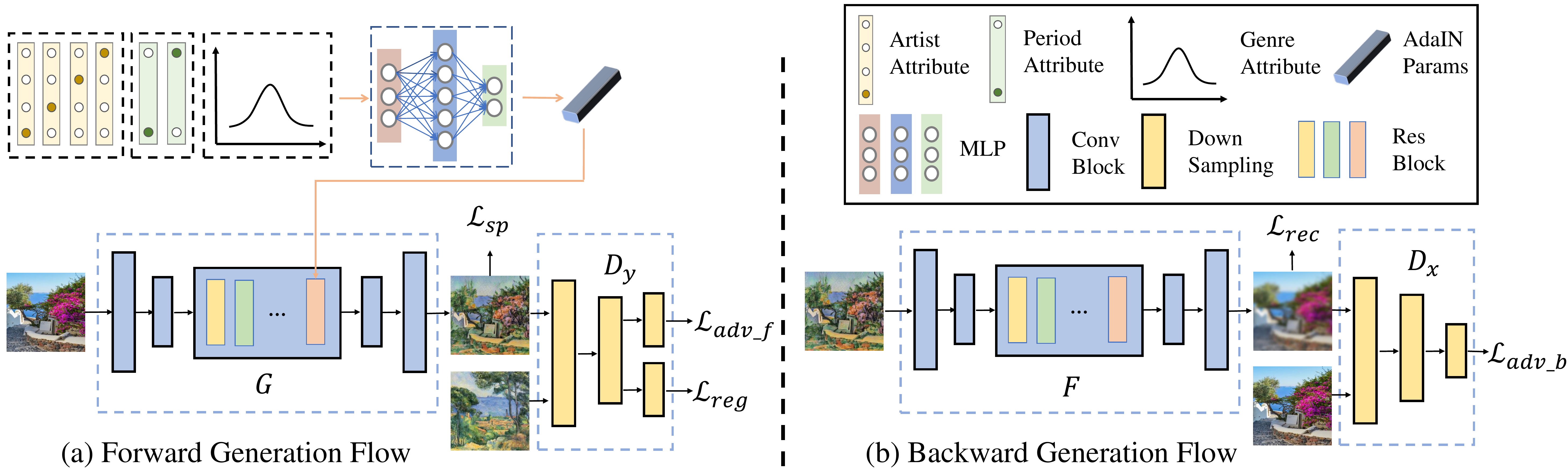}
    \caption{The asymmetric network of our proposed method. (a) A forward flow conditioned with attributes. (b) A backward synthesis stream. The legend is listed on the top right corner.}
    \label{fig:framework}
    \vspace{-3mm}
\end{figure*}
\section{Related works}
\paragraph{Painting generation}
Since Gatys et al.~\cite{Gatys:2016:ImageST} proposed convolution based style transfer method, painting generation has been widely investigated. 
Johnson et al.~\cite{Johnson:2016:PerceptualLF} adopted perception constraint and achieved the real-time synthesis of single painting. 
Sanakoyeu et al.~\cite{sanakoyeu:2018:style} accomplished multi-style generation by using a style-aware adversarial network.
Several works~\cite{Huang:2017:ICCV,park2019arbitrary,Doyle:2019:APM} emphasized the feasibility of arbitrary style transfer.
The capacity of maintaining colors and textures from random style images was still reserved without retraining.
However, the controllability of those methods is limited by exemplar-guidance based generation mechanism.
Zhu et al.~\cite{Zhu:2017:ICCV} utilized the cycle consistency to stylize unpaired images with the average domain styles. 
Choi et al.~\cite{choi2018stargan} presented a unified generator conditioned with attribute to exchange styles among multiple domains.
However, homogeneous attributes are lack of the maneuverability from various aspects.

\paragraph{Painting attributes parsing}
The analysis of intrinsic attributes of painting is challenging, some inherent properties (e.g., artist, genre and period) of artworks are extracted for deep exploration. 
Though the creation techniques change throughout painters' careers, the common style trait is still retained. 
Van Noord et al.~\cite{van:2015:toward} adopted style features from different sub-regions of painting to recognize artists. 
Taking the period factor into consideration, Yang et al.~\cite{yang:2018:historical} discovered that paintings' distribution organized according to style.
These attributes carry enough style information.

\section{Framework}

\begin{figure*}
\centering
\includegraphics[width=\linewidth]{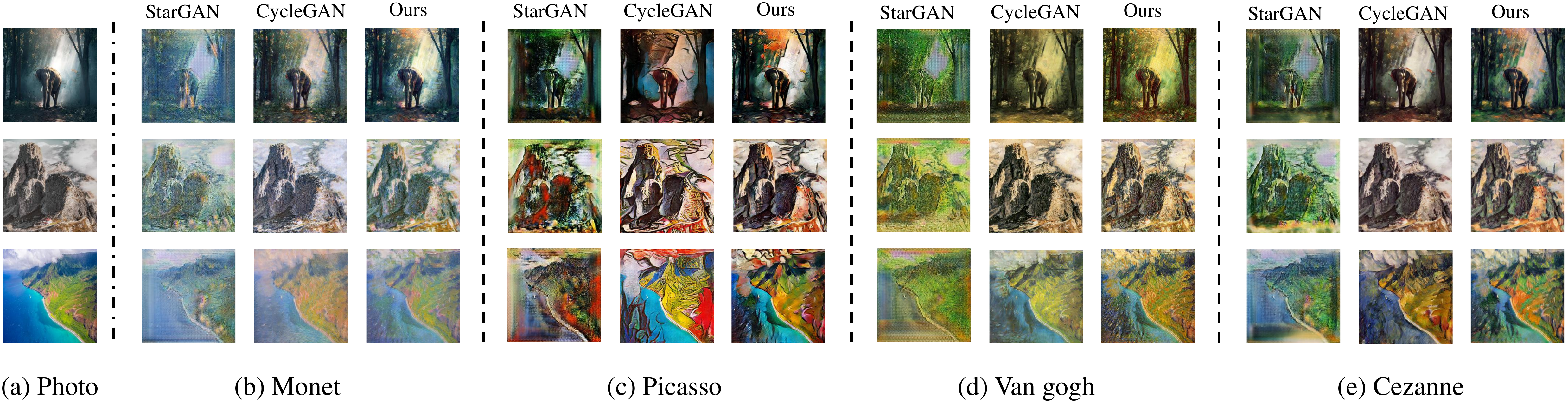}
\caption{Qualitative comparison. (a) The input images. (b)-(e) Stylized results from different models (from left to right): StarGAN~\cite{choi2018stargan}, CycleGAN~\cite{Zhu:2017:ICCV} and ours.}
\vspace{-4mm}
\label{fig:exp_comparison}
\end{figure*}

Assuming \belongto{\contentimage}{\contentdomain} is an image from \emph{content domain} and the attribute set \mathenv{\attribute=(\artist,\period,\genre)} represents three natural properties of paintings: \textit{Artist, Period} and \textit{Genre}. 
The goal of model is to generate the stylized images \mathenv{\stylizedimage}, carrying the corresponding characteristics for specific attribute set \mathenv{\attribute}. 
To this end, as shown in Figure~\ref{fig:framework}, we propose an asymmetric cycle synthesis framework composed of two unidirectional generators \mathenv{\forwardgen} and \mathenv{\backwardgen} to make controllable painting generation. 
Meanwhile, two inconsistent discriminators \mathenv{\forwarddis} and \mathenv{\backwarddis} take responsibility to distinguish the real and fake images and regress the style attributes. 

\paragraph{Conditional forward synthesis} Figure~\ref{fig:framework}(a) depicts the forward generation flow.
The content image \mathenv{\contentimage} is fed into the multiple convolution blocks for downsampling, then several residual blocks are used to encode the content feature maps. 
The attributes of artist \mathenv{\artist} and period \mathenv{\period} are expressed as one-hot labels. 
We specially design the representation of genre \mathenv{\genre}, a small perturbation sampled from a gaussian distribution \mathenv{\gaussiandistributionwithparams} and add it to the non-zero dimension of one-hot vector as the expression of genre. 
This operation aims to introduce random variations to improve the robustness of model for constructing a smoother embedding space. 

After cascading these one-hot labels which forms the attribute aggregation \mathenv{\attribute}, we use a MLP network to parse the condition and unfolds \mathenv{\attribute} to a high dimension space as AdaIN~\cite{Huang:2017:ICCV} parameters.
The adaptive instance norm (AdaIN) technique is designed for aligning the second-order statistics of content and reference images which reflect the style.
Thus, some residual blocks are normalized by AdaIN to produce the stylized content features. Finally, instead of deconvolution, we use upsampling and convolution operations to decode the features to generate the stylized image \mathenv{\stylizedimage}. 

Based on the adversarial structure, the generated image \mathenv{\stylizedimage} and the real style image \mathenv{\styleimage} are alternately sent into a multi-task discriminator \mathenv{\forwarddis}. 
For the fidelity of stylized image, the \emph{forward adversarial loss} is used to discriminate the source of input and guide the forward generator \mathenv{\forwardgen}:
\begin{equation}
\begin{split}
\mathcal{L}_{adv\_f}&=\mathbb{E}_{y}[\mathrm{log}D_{y}(y)]\\
&+\mathbb{E}_{x,c}[\mathrm{log}(1-D_{y}(G(x,c)))],
\end{split}
\end{equation}
where the task of \mathenv{\forwarddis} is the regression of attribute \mathenv{\attribute}. 
Through estimating the conditions, the generator \mathenv{\forwardgen} can better understand the similarities and differences between styles.
The \textit{attribute regression loss} is formulated as:
\begin{equation}
\mathcal{L}_{reg}=\mathbb{E}_{y}[\mathcal{L}_{D_{y}}(c|y)]+\mathbb{E}_{x,c}[\mathcal{L}_{D_{y}}(c|G(x,c))],
\end{equation}
where \mathenv{\mathcal{L}_{D_{y}}} is the cross-entropy loss. 

To further ensure the style consistency of synthetic images and real paintings, we employ the \emph{style preserving loss} to measure the similarity. 
The feature maps of a frozen parameter VGG16 network are used to calculate the \mathenv{L1} distance of gram matrix between \mathenv{\stylizedimage} and \mathenv{\styleimage}:
\begin{equation}
\mathcal{L}_{sp}=\sum_{i=1}^{4}\lVert \mathrm{VGG}_{feat}^{i}(\stylizedimage)-\mathrm{VGG}_{feat}^{i}(\styleimage) \rVert_1.
\end{equation}

\paragraph{Asymmetric backward generation}
Unlike the structure in forward flow, without condition, the backward generator \mathenv{\backwardgen} pulls the attributes away from the \mathenv{\stylizedimage} to recover the original content image \mathenv{\contentimage}. 
In Figure~\ref{fig:framework}(b), a decoder-encoder structure constitutes \mathenv{\backwardgen}. 
We obtain the reconstructed image \mathenv{\reconsimage} by using reverse recovery.
The content of original image and the corresponding reconstruction are aligned in pixel level by the \mathenv{L1} metric.
The full \emph{reconstruction loss} is formulated as:
\begin{equation}
\begin{split}
\mathcal{L}_{rec}&=\lVert F(G(x,c))-x \rVert_1+\lVert G(F(y),c)-y \rVert_1\\
&+\lVert G(y,c)-y \rVert_1 +\lVert F(x)-x \rVert_1.
\end{split}
\end{equation}

Moreover, the \emph{backward adversarial loss} is also applied as well as the forward one. 
Without attribute regression constraint, the discriminator \mathenv{\backwarddis} only estimates the source of images:
\begin{equation}
\begin{split}
\mathcal{L}_{adv\_b}&=\mathbb{E}_{x}[\mathrm{log}D_{x}(x)]\\
&+\mathbb{E}_{y}[\mathrm{log}(1-D_{x}(F(y)))].
\end{split}
\end{equation}

\vspace{-2mm}
\paragraph{Full loss}
We combine all loss terms as the final objective:
\begin{equation}
\begin{split}
\min\limits_{\forwardgen,\backwardgen}\max\limits_{\forwarddis,\backwarddis}\mathcal{L}_{full}&=\mathcal{L}_{adv\_f}+\mathcal{L}_{adv\_b}+\lambda_{rec}\mathcal{L}_{rec}\\
&+\lambda_{reg}\mathcal{L}_{reg}+\lambda_{s}\mathcal{L}_{sp},
\end{split}
\end{equation}
where we set $\lambda_{rec} = 10, \lambda_{reg} = 1, \lambda_{s} = 1e^{-4}$ in training.

\begin{figure}
\centering
\includegraphics[width=\linewidth]{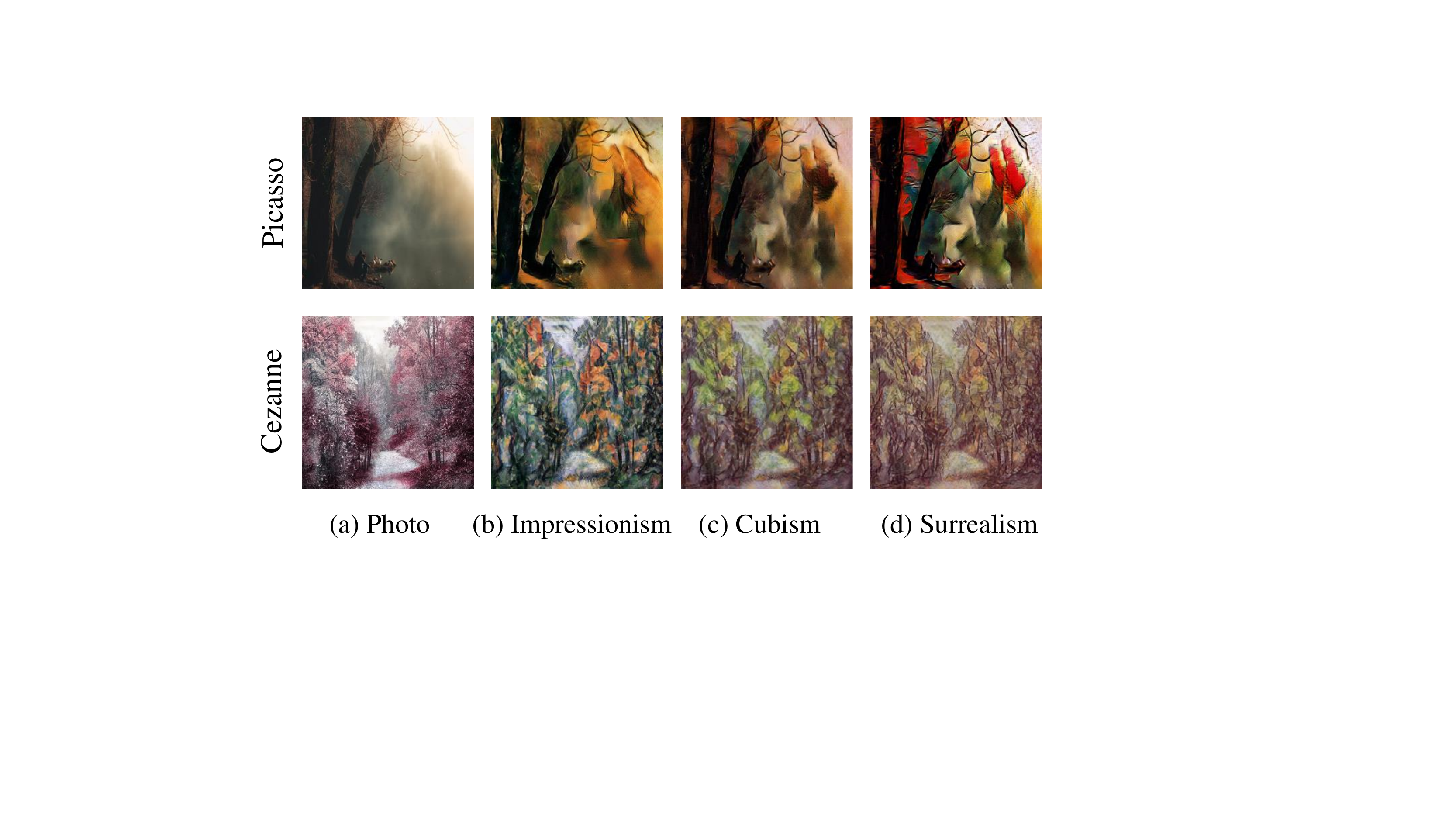}
\vspace{-3mm}
\caption{Painting generation of different genres. The results from Picasso or Van gogh are stylized with three genre attributes.}
\vspace{-2mm}
\label{fig:exp_comparison_genre}
\end{figure}
\begin{figure}
\centering
\includegraphics[width=\linewidth]{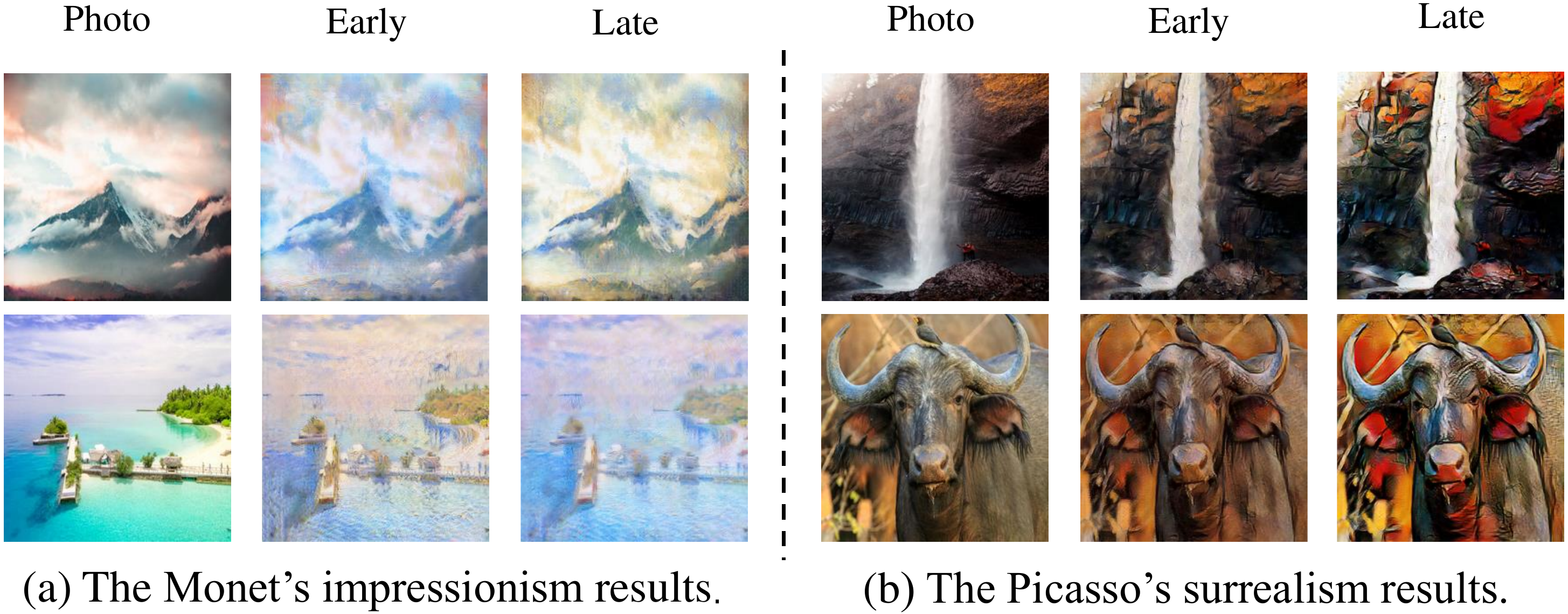}
\caption{Painting generation of different periods. The stylized images reveal the disparity between early and late periods of an artist.}
\vspace{-5mm}
\label{fig:exp_comparison_period}
\end{figure}
\section{Experiments}
\paragraph{Experimental setup}
We pick the artworks of Picasso, C\'ezanne, Monet and Van Gogh as traning data to test the algorithm.
The artworks and corresponding painting attributes are downloaded from \emph{Wikiart.com}. 
For each artist, we select two representative periods which appear obvious changes in the art history, e.g., \emph{Cubist} and \emph{Neoclassicist} period of Picasso, \emph{Mature} and \emph{Final} period of C\'ezanne. 
If an artist's works are lack of clear period definition, we just reserve paintings in the early and late time of his/her career. 
For genres, \emph{Impressionism}, \emph{Cubism} and \emph{Surrealism} are used. 
Content images are collected from \emph{Pexels.com}. 
The resolution of our input and output images is $256\times256$.

\vspace{-2mm}
\paragraph{Qualitative evaluation}
In Figure~\ref{fig:exp_comparison}, the quality of images stylized by StarGAN~\cite{choi2018stargan}, CycleGAN~\cite{Zhu:2017:ICCV} and ours are compared under the control of artist attributes. 
We retrain these models on our dataset.
The period and genre attributes of our model are fixed for a fair comparison. 
We can see that StarGAN easily confuses the styles between domains so that the results are affected by other artists' style. 
Due to the adoption of deconvolution, CycleGAN produces more artifacts. 
Our method successfully perserves the variations among domains and obtains high quality results.

Figure~\ref{fig:exp_comparison_genre} demonstrates the controllability of our method for genres.
Conditioned by artist and period, the textures and colors of stylized images change with genres. 
We use fixed one-hot vector for genre in test. 
The main characteristics of genre are extracted (e.g., the green fits with the habit of C$\acute{e}$zanne in impressionism) and are expressed well. 
Particularly, we also create some zero-shot results (e.g., C$\acute{e}$zanne's Surrealism) by mixing the attributes. 
Figure~\ref{fig:exp_comparison_period} displays the influence of period.
Considering the brush stroke strongly associating with the unaltered genre, colors become the major variations. 
For instance, the Picasso's surrealism results in late period are more bright and abstract than the early ones.

\vspace{-2mm}
\paragraph{Quantitative evaluation}
We calculate the artist classification accuracy and the IS metric for StarGAN and our method, which both contain a unified generator for a fair comparison. 
The accuracy computed by a finetuned ResNet-18 network examines the similarity between stylized images and artist domains. 
The IS metric indicates the reality of stylized images, testing on a finetuned Inception-V3 network.
As shown in Table~\ref{tab:accuracy_is}, our method can outperform the StarGAN method~\cite{choi2018stargan}.
\begin{table}
\centering
\resizebox{.49\textwidth}{!}{
\begin{tabular}{|c|c|c|c|c|c|c|c|c|c|c|c|c|}
\hline
\multirow{2}{*}{Method} &
\multicolumn{2}{c|}{Photo2Monet} & 
\multicolumn{2}{c|}{Photo2Picasso} &
\multicolumn{2}{c|}{Photo2Vangogh} & 
\multicolumn{2}{c|}{Photo2C$\acute{e}$zanne} \\
\cline{2-9}
  & IS & Acc & IS & Acc & IS & Acc & IS & Acc\\
\hline
StarGAN & 1.53 & 0.45 & 1.37 & 0.07  & 1.17 & 0.87 & 1.59 & 0.12 \\
\hline
Ours & 1.30 & 0.90 & 1.87 & 0.62  & 1.52 & 0.73 & 2.23 & 0.55 \\
\hline
\end{tabular}
}
\caption{The artist classification accuracy and IS score of methods with a unified generator.}
\label{tab:accuracy_is}
\vspace{-3mm}
\end{table}

\section{Conclusions}
We propose a multi-attribute guided stylization method to increase more controllability for painting generation.
Internal painting properties (e.g., artist, genre and period) are utilized as conditions. 
An asymmetrical cycle structure with control branch constitutes our gframework. 
The qualitative and quantitative evaluations show the superiority of our model over the state-of-the-art methods.
\section*{Acknowledgements}
This work was supported by National Natural Science Foundation of China under nos. 61832016 and 61672520.

\bibliographystyle{latex8}
\bibliography{PainterGAN}

\end{document}